\documentclass[11pt]{article}

\usepackage[T1]{fontenc}
\usepackage[utf8]{inputenc}
\usepackage{lmodern}
\usepackage[margin=1in]{geometry}
\usepackage{microtype}
\usepackage{booktabs}
\usepackage{tabularx}
\usepackage{longtable}
\usepackage{array}
\usepackage{multirow}
\usepackage{amsmath}
\usepackage{amssymb}
\usepackage{graphicx}
\usepackage{tikz}
\usetikzlibrary{arrows.meta,positioning,shapes.geometric}
\usepackage{natbib}
\usepackage[hidelinks]{hyperref}
\usepackage{url}
\usepackage{xcolor}
\usepackage{enumitem}
\usepackage{caption}
\usepackage{threeparttable}

\title{\textbf{Scaling Down the Scaling Laws: Parameter Efficiency and Compute-Optimal Training in Resource-Constrained Large Language Models}}
\author{Joe Dwyer\\
\small ECPI University\\
\small jdwyer@ecpi.edu}
\date{}

\begin{document}
\maketitle

\begin{abstract}
Large language model (LLM) development has been shaped by a simple but increasingly costly premise: improvements in model capability can be obtained by scaling parameters, training data, and computational resources. The empirical scaling laws formalized by \citet{kaplan2020} established predictable relationships among model size, dataset size, compute, and loss, while \citet{hoffmann2022} subsequently showed that many large models were undertrained relative to their parameter counts and that compute-optimal training requires a more deliberate balance between tokens and parameters. As training costs, infrastructure requirements, and environmental burdens have increased, the research question has therefore shifted from how far LLMs can be scaled to how efficiently available compute can be allocated.

This structured narrative review synthesizes literature on neural scaling laws, token utilization, parameter efficiency, compute overhead, optimized training strategies, and LLM operation in resource-constrained and edge environments. The review argues that the scaling literature has progressed through three overlapping stages: scale maximization, compute-optimal balancing, and resource-constrained optimization. Across these stages, a common pattern emerges: efficiency is not determined by model size alone but by the interaction of data quantity and quality, parameter count, architecture, training procedure, memory and communication overhead, and hardware constraints. The literature also reveals three persistent gaps. First, scaling-law results derived on enterprise-grade infrastructure have not been sufficiently tested under constrained compute. Second, parameter efficiency lacks a standardized operational definition that combines model quality with both model capacity and compute cost. Third, resource-conscious LLM research remains methodologically fragmented, limiting reproducibility and direct comparison across studies. These gaps motivate a research agenda centered on controlled token-to-parameter experimentation, energy-aware efficiency metrics, reproducible constrained-compute benchmarks, and explicit tests of whether compute-optimal scaling principles generalize downward to smaller hardware environments.
\end{abstract}


\section{Introduction}

The modern development of large language models has been inseparable from scaling. Larger models, larger datasets, and larger compute budgets produced substantial improvements in language modeling and helped transform natural language processing (NLP) from a collection of task-specific techniques into a foundation for broadly capable pretrained models \citep{brown2020,kaplan2020,radford2019}. Yet the same trajectory that produced capability gains also intensified the computational and economic demands of model development. Training frontier-scale systems increasingly requires specialized hardware, substantial energy consumption, and capital resources unavailable to many smaller organizations and independent researchers \citep{cottier2024,hutong2024}. The central problem is therefore no longer merely whether additional scale improves performance. It is whether additional scale remains an efficient use of finite computational resources.

The scaling-law literature provides the theoretical and empirical foundation for this shift. \citet{kaplan2020} demonstrated that language-model loss follows predictable power-law relationships with model size, dataset size, and compute. Their results established scaling as a measurable phenomenon rather than an informal engineering intuition. However, power-law behavior also implies diminishing returns: improvements continue as resources increase, but the marginal gain obtained from additional scale decreases. \citet{hoffmann2022} refined this picture by showing that many large models had been trained with too few tokens relative to their parameter counts. Their compute-optimal analysis suggested that model size and training data should be jointly balanced under a fixed compute budget rather than increased independently.

This refinement changed the practical interpretation of scaling. Under a compute-optimal view, the central question is not simply whether a larger model can be trained, but whether a given allocation of parameters, data, and compute extracts the greatest useful performance from the available resource budget. That change in emphasis is especially important for resource-constrained research. If effective training depends on allocation rather than scale alone, then smaller models and smaller hardware environments may provide meaningful experimental platforms for studying efficiency, provided that token volume, parameter count, architecture, and compute are controlled appropriately.

A second line of research reinforces this transition. Data-pruning methods emphasize data quality rather than indiscriminate quantity \citep{sorscher2022}; scale-time equivalence questions the assumption that larger architectures are always necessary \citep{boopathy2024}; efficient transformer research investigates sparse and mixture-of-experts architectures \citep{tay2022}; and systems research targets memory, communication, quantization, and low-resource deployment \citep{narayanan2021,dettmers2023,li2024,yi2023,yu2024}. Together, these studies suggest that the field is moving from a scale-maximization paradigm toward a multidimensional efficiency paradigm.

The purpose of this review is to synthesize that transition with particular attention to parameter efficiency and constrained compute. Rather than treating scaling laws, token utilization, architectural optimization, systems efficiency, and edge computing as separate literatures, the review organizes them around a common question: \emph{How should limited compute be allocated to obtain useful LLM performance?} The paper makes four contributions. First, it traces the intellectual transition from empirical scaling laws to compute-optimal training. Second, it integrates token utilization, parameter efficiency, and compute overhead into a common efficiency framework. Third, it organizes resource-efficient LLM methods into a five-part taxonomy spanning data, parameters, architecture, systems, and hardware. Fourth, it identifies empirical, theoretical, and methodological gaps that arise when scaling principles developed on large infrastructure are applied to constrained environments.

\section{Review Scope and Search Strategy}

This paper is a structured narrative review refactored from a dissertation literature review. The source corpus was assembled through searches of IEEE Xplore, Google Scholar, the ACM Digital Library, and arXiv. The dissertation search process emphasized work published from 2021 through 2025 while also incorporating older foundational sources when necessary to establish the historical and theoretical lineage of scaling research. Because the original review was designed to support an experimental dissertation rather than a formal systematic review, the search process should be interpreted as structured and reproducible in intent, but not as a PRISMA-style systematic review.

Table~\ref{tab:searches} reproduces the principal search logic and result counts documented in the dissertation. Several search categories were intentionally broad because the review sought to connect LLM training efficiency with methodological and hardware constraints. The edge-computing query, for example, initially returned a very large result set and was later refined to focus on edge devices or edge hardware in combination with LLM training and efficiency terms. Searches for direct combinations of LLM training costs with financial or environmental impact produced no relevant results in the documented search pass, requiring the review to rely on adjacent literature concerning training costs, transparency, data-center expansion, and accessibility.

\begin{table}[htbp]
\centering
\caption{Documented Search Queries in the Source Literature Review}
\label{tab:searches}
\small
\begin{tabularx}{\textwidth}{@{}Xrrl@{}}
\toprule
\textbf{Search term} & \textbf{Total results} & \textbf{Relevant studies} & \textbf{Date range} \\
\midrule
``LLM Training'' AND (``Compute Overhead'' OR ``Resource Utilization'') & 371 & 90 & 2021--2025 \\
(``Parameter Efficiency'' OR ``Token Optimization'') AND ``Scaling Laws'' & 333 & 37 & 2021--2025 \\
``Compute-Optimal Scaling'' AND (``Hoffmann et al.'' OR ``Scaling Efficiency'') & 77 & 1 & 2021--2025 \\
``Repeated Measures Design'' AND (``AI Experiments'' OR ``Machine Learning'') & 1101 & 15 & 2021--2025 \\
``Edge Computing'' AND (``Model Training'' OR ``Resource Constraints'') & 19872 & 15 & 2021--2025 \\
``Large Language Models'' AND ``Compute Efficiency'' AND ``Training Costs'' & 44 & 19 & 2021--2025 \\
``LLM Training Costs'' AND (``Financial Impact'' OR ``Environmental Footprint'') & 0 & 0 & 2021--2025 \\
``Kaplan Scaling Laws'' AND ``Neural Language Models'' & 33 & 15 & 2021--2025 \\
\bottomrule
\end{tabularx}
\vspace{0.5em}
\begin{minipage}{0.96\textwidth}
\footnotesize
\textit{Note.} The initial edge-computing query was subsequently refined to (``Edge Device'' OR ``Edge Hardware'') AND ``LLM Training'' AND (``Parameter Efficiency'' OR ``Compute Optimization''), reducing the result set to 142 in the source review.
\end{minipage}
\end{table}

The present synthesis prioritizes six recurring themes: (1) the historical transition from statistical and neural NLP to transformer-based language models; (2) neural scaling laws and compute-optimal refinements; (3) parameter efficiency and token utilization; (4) compute overhead and resource-efficient training methods; (5) low-power and edge LLM research; and (6) accessibility, reproducibility, and sustainability. The review also distinguishes foundational scaling theory from engineering techniques that reduce the cost of implementing or adapting LLMs.

\section{From NLP Scaling to LLM Scaling}

The concern with efficiency predates modern LLMs. Early NLP systems such as ELIZA relied on rule-based pattern matching and shallow representations of local context \citep{weizenbaum1966,cambria2014}. Statistical approaches later displaced many hand-written systems, with n-gram and Hidden Markov methods establishing a more data-driven paradigm \citep{russell2021}. By the 2010s, distributed word representations such as Word2Vec and GloVe encoded semantic relationships in dense vector spaces and helped complete the transition from statistical NLP to neural NLP \citep{mikolov2013,pennington2014,goldberg2016}.

Static embeddings nevertheless assigned the same representation to a word regardless of surrounding context. Contextualized representations, including ELMo, addressed this limitation by generating token representations conditioned on neighboring language \citep{peters2018}. Transformer architectures then provided a scalable mechanism for modeling long-range dependencies through attention \citep{vaswani2017}. BERT combined transformer-based bidirectional context with large-scale pretraining and task-specific fine-tuning \citep{devlin2019}, while the GPT line demonstrated the effectiveness of autoregressive pretraining and increasingly large parameter counts \citep{radford2018,radford2019,brown2020}.

This historical progression matters to a review of compute efficiency because each improvement in representational power also altered the computational regime. Transformer-based pretraining created a path toward increasingly general models, but it also made the interaction among parameters, tokens, and compute a first-order research problem. The success of large autoregressive models therefore created both the empirical motivation and the experimental scale necessary for formal scaling laws to emerge.

\section{Scaling Laws and the Compute-Optimal Turn}

\subsection{Kaplan et al.: Scaling as a Predictable Empirical Relationship}

\citet{kaplan2020} provided the foundational empirical account of neural language-model scaling. Their central contribution was to show that model performance, particularly loss, changes predictably as model size, dataset size, and compute increase. This result made it possible to reason about model growth using quantitative relationships rather than ad hoc assumptions. The practical implication was powerful: if larger models and larger datasets follow stable trends, then performance at one scale may inform expectations at another.

The same framework, however, also formalized diminishing returns. Power-law improvement does not imply constant marginal benefit. As resources grow, each additional unit of parameters, data, or compute produces a smaller improvement in loss. This observation is critical for constrained environments because it implies that inefficient resource allocation can become increasingly expensive at larger scales. Scale remains productive, but scale alone is not an optimization criterion.

The dissertation from which this review is derived drew a conceptual parallel between this limiting behavior and classical asymptotic analysis \citep{knuth1997}. That parallel is best treated as a conceptual lens rather than an equivalence. Big-O analysis characterizes algorithmic growth independent of a specific hardware platform, whereas empirical scaling laws characterize observed relationships among model performance and training resources. Both, however, encourage reasoning about behavior as inputs and resource demands increase. For LLMs, that analogy foregrounds the distinction between raw growth and efficient growth.

\subsection{Hoffmann et al.: Balancing Tokens and Parameters}

\citet{hoffmann2022} substantially revised the operational interpretation of scaling. Their analysis of compute-optimal training showed that many large models were undertrained relative to their parameter counts. Under a fixed compute budget, increasing the number of parameters without allocating a commensurate amount of training data could produce a model that was larger but not compute-optimal. Their Chinchilla results demonstrated that a smaller model trained on substantially more tokens could outperform larger alternatives under comparable compute constraints.

This result shifted the field from a model-size-centered view toward a joint optimization problem. Parameters and training tokens are not independent levers. They compete for the same compute budget. A compute-optimal strategy must therefore determine how much of that budget should be allocated to model capacity and how much should be allocated to learning from data. The dissertation summarized this relationship using the proportionality

\begin{equation}
D \propto N^{0.73},
\end{equation}

where $D$ denotes training data and $N$ denotes model parameters as represented in the source review of Hoffmann et al. This relationship was used in the dissertation as a practical expression of sublinear token scaling and as a basis for comparing earlier and later LLM training strategies. Because scaling-law estimates are model- and methodology-dependent, the exact exponent should not be treated as a universal constant; the broader point is the requirement to balance model size and data under finite compute.

\subsection{Refinements and Challenges to Simple Scaling}

Subsequent studies broadened the scaling-law literature in several directions. \citet{bahri2021} investigated theoretical explanations for observed neural scaling behavior using statistical and geometric learning principles, while \citet{hutter2021} approached learning curves from a more formal theoretical perspective. These works differ methodologically from the empirical orientation of Kaplan and Hoffmann but reinforce the premise that performance trends can exhibit regular structure as resources increase.

Other studies question whether model size should retain its privileged position in scaling discussions. \citet{boopathy2024} explored scale-time equivalence, suggesting that smaller networks trained longer can under some conditions achieve performance comparable to larger models. \citet{sorscher2022} showed that data pruning and data quality can alter expected scaling behavior, demonstrating that intelligently selected data may outperform naive increases in dataset size. \citet{sengupta2025} surveyed conditions under which neural scaling laws hold or break down and emphasized architecture- and context-dependent limitations. \citet{wang2024} further examined scaling behavior across architectural variants, reinforcing the importance of architecture when generalizing scaling relationships.

Table~\ref{tab:theories} summarizes the theoretical landscape represented in the source review.

\begin{table}[htbp]
\centering
\caption{Selected Scaling Frameworks and Their Relevance to Compute Efficiency}
\label{tab:theories}
\small
\begin{tabularx}{\textwidth}{@{}p{0.20\textwidth}p{0.22\textwidth}Xp{0.25\textwidth}@{}}
\toprule
\textbf{Study} & \textbf{Primary focus} & \textbf{Key insight} & \textbf{Efficiency relevance} \\
\midrule
\citet{kaplan2020} & Empirical scaling laws & Loss scales predictably with model size, data, and compute. & Establishes the baseline relationship among resources and performance. \\
\citet{bahri2021} & Theory of scaling laws & Statistical and geometric principles help explain observed power laws. & Adds theoretical structure to empirical scaling behavior. \\
\citet{hutter2021} & Learning-curve theory & Formal learning bounds provide an alternative perspective on performance growth. & Frames efficiency and learning limits theoretically. \\
\citet{hoffmann2022} & Compute-optimal scaling & Model size and training tokens should be balanced under fixed compute. & Reorients scaling around token-to-parameter allocation. \\
\citet{boopathy2024} & Scale-time equivalence & Smaller models trained longer can sometimes approach larger-model performance. & Challenges model-size dominance. \\
\citet{sorscher2022} & Data-centric pruning & Data quality and selection can improve scaling efficiency beyond naive data growth. & Shifts attention from data quantity to data utility. \\
\citet{sengupta2025} & Scaling-law survey & Scaling behavior depends on architecture, context, and methodological assumptions. & Highlights limits on generalizing a single scaling rule. \\
\bottomrule
\end{tabularx}
\end{table}

Taken together, this literature changes the meaning of scaling. Scaling is no longer best understood as the independent growth of parameters, data, and compute. It is increasingly understood as a constrained optimization problem in which the usefulness of additional resources depends on how those resources are allocated.

\section{Parameter Efficiency and Token Utilization}

Parameter efficiency provides a natural bridge between scaling theory and resource-constrained practice. In broad terms, parameter efficiency concerns how effectively a model converts its parameter capacity and available compute into useful performance. Token utilization concerns how effectively the training process converts available tokens into learned predictive capability. These concepts are related because the usefulness of a given parameter count depends in part on how much and what kind of data are used to train those parameters.

The compute-optimal results of \citet{hoffmann2022} make this connection explicit. A large parameter count trained on too few tokens can be inefficient because model capacity is not fully utilized. Conversely, increasing tokens indefinitely is also not guaranteed to be efficient because training cost rises with token volume and the marginal value of additional data may decline. The relevant optimization target is therefore not maximum token count or maximum parameter count, but an appropriate relationship between the two under the available compute budget.

\citet{maloney2022} provided a solvable statistical account of neural scaling behavior in which model performance depends jointly on parameter count and dataset size. Their theoretical treatment complements the empirical findings of Kaplan and Hoffmann by reinforcing the idea that neither model capacity nor data scale should be analyzed in isolation. Data-centric work provides another qualification: if the effective value of tokens differs by quality, then token count alone may be an incomplete measure of useful training exposure \citep{sorscher2022,pang2024}.

A central difficulty is that \emph{parameter efficiency} is not yet a standardized metric across the literature reviewed here. Studies variously report loss, perplexity, accuracy, throughput, memory use, training time, or performance under a fixed compute budget. The source dissertation operationalized parameter efficiency using inverse perplexity divided by the product of model parameters and compute throughput:

\begin{equation}
\mathrm{PE} = \frac{1/\mathrm{PPL}}{N \times \mathrm{TFLOPS}},
\label{eq:pe}
\end{equation}

where $\mathrm{PPL}$ is perplexity, $N$ is the number of model parameters, and TFLOPS represents computational throughput. Equation~\ref{eq:pe} should be understood here as one concrete operationalization rather than an established field-wide standard. Its conceptual value is that it makes the denominator of performance explicit: model quality is evaluated relative to both model capacity and compute. The absence of consensus around such a metric is itself an important research gap.

This gap becomes especially consequential when studies are compared across hardware environments. A method that reduces perplexity but doubles compute demand may be desirable for one objective and inefficient for another. Likewise, a smaller model may be attractive because of memory or deployment constraints even if it requires longer training. A resource-conscious literature therefore requires metrics that make trade-offs visible rather than reducing efficiency to a single performance outcome.

\section{Compute Overhead as a Multilayer Constraint}

Compute overhead is a useful organizing concept because inefficient LLM training can arise at multiple layers of the stack. In general computing, overhead refers to resources consumed beyond the work directly necessary to produce the target result. Examples include communication delays in parallel systems and redundant memory operations in software \citep{mele2023,su2019}. In LLM training, analogous overhead appears in data processing, optimization, synchronization, memory movement, communication, and underutilized hardware.

At the model-training level, compute-optimal scaling addresses one important form of overhead: allocating too much compute to model capacity and too little to training data, or vice versa \citep{hoffmann2022}. At the systems level, distributed training may lose efficiency to synchronization and communication. \citet{narayanan2021}, for example, studied large-scale language-model training using Megatron-LM and DeepSpeed with an emphasis on memory and compute optimization across GPU clusters. At the representation level, quantization reduces the precision required to store and operate on model weights and activations, creating another route to lower resource demand \citep{dettmers2023}. At the architecture level, sparse and mixture-of-experts approaches seek to increase effective capacity without activating all parameters for every token \citep{tay2022}.

These examples demonstrate why compute efficiency cannot be reduced to one intervention. A training run may be data-inefficient, parameter-inefficient, memory-inefficient, communication-inefficient, or hardware-inefficient. Improvements at one layer can also expose bottlenecks at another. Reducing model precision may lower memory pressure but leave data loading unchanged; pruning data may reduce training steps but not improve communication efficiency; increasing token volume may improve utilization of model capacity while increasing total energy consumption. Efficiency is therefore best understood as a system property.

\begin{figure}[htbp]
\centering
\begin{tikzpicture}[
    node distance=1.25cm and 1.0cm,
    box/.style={draw, rounded corners, align=center, minimum width=3.5cm, minimum height=1.15cm, text width=3.2cm},
    smallbox/.style={draw, rounded corners, align=center, minimum width=2.75cm, minimum height=0.95cm, text width=2.5cm},
    arrow/.style={-{Latex[length=3mm]}, thick}
]
\node[box] (scale) {\textbf{Scale Maximization}\\Increase parameters, data, and compute};
\node[box, right=of scale] (optimal) {\textbf{Compute-Optimal Balancing}\\Allocate tokens and parameters under fixed compute};
\node[box, right=of optimal] (constrained) {\textbf{Resource-Constrained Optimization}\\Optimize quality, architecture, systems, and hardware};

\draw[arrow] (scale) -- (optimal);
\draw[arrow] (optimal) -- (constrained);

\node[smallbox, below=1.15cm of constrained, xshift=-3.2cm] (data) {Data / Token\\Efficiency};
\node[smallbox, below=1.15cm of constrained] (param) {Parameter\\Efficiency};
\node[smallbox, below=1.15cm of constrained, xshift=3.2cm] (arch) {Architecture\\Efficiency};
\node[smallbox, below=1.15cm of param, xshift=-1.6cm] (sys) {Systems\\Efficiency};
\node[smallbox, below=1.15cm of param, xshift=1.6cm] (hw) {Hardware / Edge\\Efficiency};

\draw[arrow] (constrained.south) -- ++(0,-0.45) -| (data.north);
\draw[arrow] (constrained.south) -- (param.north);
\draw[arrow] (constrained.south) -- ++(0,-0.45) -| (arch.north);
\draw[arrow] (param.south) -- ++(0,-0.35) -| (sys.north);
\draw[arrow] (param.south) -- ++(0,-0.35) -| (hw.north);
\end{tikzpicture}
\caption{A synthesis of the shift from scale maximization to compute-optimal and resource-constrained LLM research. The five efficiency dimensions are derived from the themes represented in the source literature review.}
\label{fig:transition}
\end{figure}
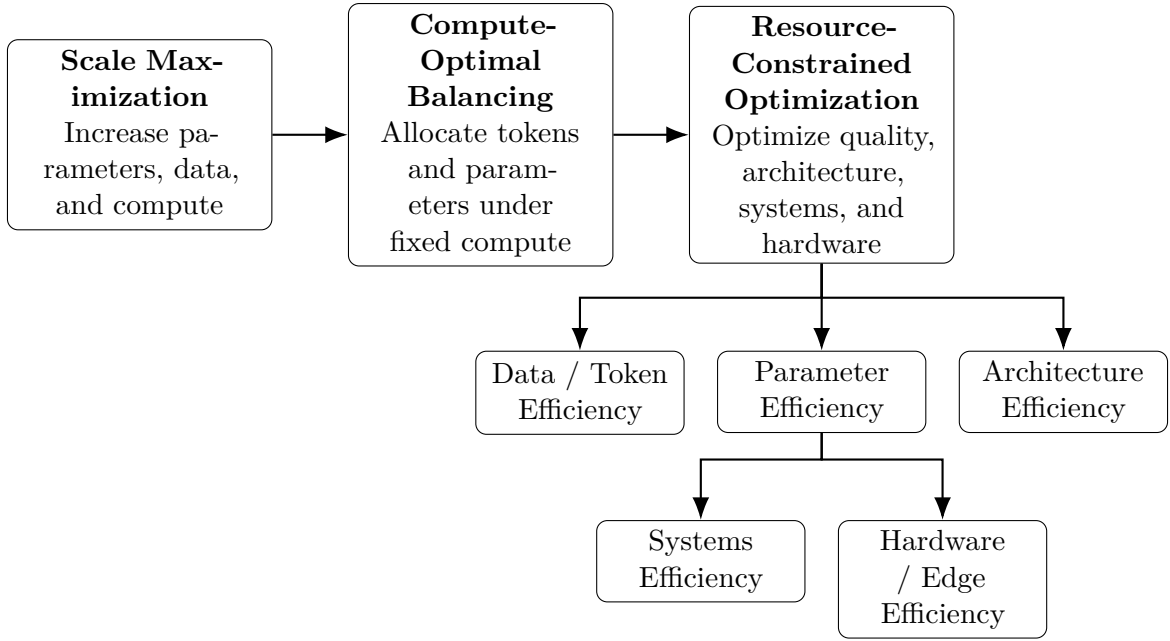

Figure~\ref{fig:transition} synthesizes the progression represented in the reviewed literature. The figure does not claim that the stages are historically discrete; rather, it illustrates a change in emphasis from adding resources, to balancing resources, to optimizing within explicit resource constraints.

\section{From Traditional Scaling to Compute-Optimized Training}

The contrast between traditional and compute-optimized training is visible in the parameter and token configurations of influential models. Early large-scale language models generally increased model size aggressively as compute became available. GPT-2, GPT-3, Gopher, Jurassic-1, Megatron-Turing NLG, and PaLM exemplify the period in which increasing parameter count and dataset scale were central strategies \citep{radford2019,brown2020,rae2021,lieber2021,smith2022,chowdhery2022}.

Table~\ref{tab:traditional} reproduces the comparative values documented in the dissertation. The values should be treated as historical approximations because reporting practices vary across model releases and some token counts are estimates.

\begin{table}[htbp]
\centering
\caption{Examples of Models Associated with Pre-Compute-Optimal Scaling Practices}
\label{tab:traditional}
\small
\begin{tabular}{@{}llll@{}}
\toprule
\textbf{Model} & \textbf{Developer} & \textbf{Parameters} & \textbf{Estimated tokens} \\
\midrule
GPT-2 & OpenAI & 1.5B & $\sim$40B \\
GPT-3 & OpenAI & 175B & $\sim$300B \\
Gopher & DeepMind & 280B & $\sim$300B \\
MT-NLG & Microsoft/NVIDIA & 530B & $\sim$270--400B \\
Jurassic-1 & AI21 Labs & 178B & $\sim$300B \\
PaLM & Google & 540B & $\sim$780B \\
Megatron-Turing NLG & Microsoft/NVIDIA & 530B & $\sim$500B \\
\bottomrule
\end{tabular}
\vspace{0.5em}

\begin{minipage}{0.94\textwidth}
\footnotesize
\textit{Sources:} \citet{radford2019}, \citet{brown2020}, \citet{rae2021}, \citet{smith2022}, \citet{lieber2021}, and \citet{chowdhery2022}.
\end{minipage}
\end{table}

The significance of \citet{hoffmann2022} was not that larger models ceased to be useful, but that the allocation of compute became the relevant optimization target. Chinchilla demonstrated that a substantially smaller model could outperform larger alternatives when trained on more data under a comparable compute budget. Later small and mid-scale models also reflected growing interest in training efficiency, open experimentation, and improved data utilization. The dissertation highlighted Phi-1, Phi-2, Gemma, Mistral, and TinyLlama as examples relevant to this shift \citep{gunasekar2023,javaheripi2023,mesnard2024,jiang2023,zhang2024}. Table~\ref{tab:optimized} now reports only training-token counts that could be supported by the primary model papers or official model documentation used in this quality-control pass.

\begin{table}[htbp]
\centering
\caption{Examples Used in the Source Review to Illustrate Compute-Optimized or Resource-Conscious Training}
\label{tab:optimized}
\small
\begin{threeparttable}
\begin{tabular}{@{}llll@{}}
\toprule
\textbf{Model} & \textbf{Developer} & \textbf{Parameters} & \textbf{Training tokens} \\
\midrule
Chinchilla & DeepMind & 70B & 1.4T \\
Phi-1 & Microsoft & 1.3B & 54B$^{a}$ \\
Phi-2 & Microsoft & 2.7B & 1.4T \\
Gemma 2B & Google & 2B & 2T \\
Gemma 7B & Google & 7B & 6T \\
TinyLlama & Open source & 1.1B & $\sim$1T \\
\bottomrule
\end{tabular}
\begin{tablenotes}[flushleft]
\footnotesize
\item[$a$] The Phi-1 model card reports 54B training tokens drawn from 7B unique tokens. Values in this table are limited to counts supported by primary model papers or official model documentation; the earlier dissertation estimates for Gemma 9B/27B and Mistral 7B are omitted here because they were not supported by the cited primary sources.
\end{tablenotes}
\end{threeparttable}
\end{table}

TinyLlama occupies a different role in this literature. It is not itself a scaling-law proposal, but its small 1.1B-parameter architecture, public training artifacts, and permissive research context make it useful for experiments that would be impractical with frontier-scale systems \citep{zhang2024}. The broader importance of such models is methodological: they permit researchers to test hypotheses about token allocation, data quality, adaptation, and compute efficiency at a scale where repeated experimentation is financially and technically feasible.

\section{A Taxonomy of Resource-Efficient LLM Training}

The reviewed literature can be organized into five interacting forms of efficiency. This taxonomy is useful because many papers described as ``efficient LLM'' research optimize different denominators. Some reduce tokens, some reduce active parameters, some reduce memory, some reduce communication, and others target constrained hardware. Treating these as equivalent obscures the trade-offs that matter in practice.

\subsection{Data and Token Efficiency}

Data and token efficiency concern the amount and quality of training information required to achieve a target level of performance. Compute-optimal scaling makes token allocation central because undertraining wastes model capacity \citep{hoffmann2022}. Data pruning further complicates the relationship by showing that the value of examples is not uniform \citep{sorscher2022}. Work on data curation and preference-driven optimization similarly treats training data as a resource whose selection can affect downstream compute requirements \citep{bernardelle2024,pang2024}.

For constrained researchers, token efficiency is attractive because it is one of the few scaling variables that can be manipulated without changing model architecture or hardware. However, token count is also an imperfect proxy for information content. Two datasets with equal token counts can differ substantially in diversity, redundancy, quality, and task relevance. Future efficiency studies therefore need to report both token quantity and data-selection procedures.

\subsection{Parameter Efficiency}

Parameter efficiency asks how much useful performance is obtained from a given model capacity. Compute-optimal scaling shows that parameters can be wasted when not supported by adequate training data \citep{hoffmann2022}. Scale-time equivalence further suggests that longer training of smaller models can sometimes substitute for additional parameters \citep{boopathy2024}. Parameter-efficient adaptation techniques, including low-rank and quantized approaches, represent a related but distinct branch of the literature because they reduce the number or precision of trainable parameters during adaptation rather than pretraining an entire model from scratch \citep{dettmers2023}.

A publication-quality efficiency framework should distinguish at least three questions: how efficiently parameters are used during pretraining, how efficiently they are updated during adaptation, and how efficiently they are stored and executed during inference. Conflating these questions can make two ``parameter-efficient'' methods appear directly comparable when they optimize different stages of the model lifecycle.

\subsection{Architectural Efficiency}

Architectural efficiency changes the structure of the model so that performance can improve without a proportional increase in active computation. Sparse models and mixture-of-experts architectures are important examples because they can expand nominal capacity while activating only portions of the network for a given input \citep{tay2022}. Architectural research therefore weakens any assumption that parameter count alone captures effective training demand.

Scaling-law generalization becomes more difficult as architectures diverge. If dense transformers, sparse models, and mixture-of-experts systems allocate computation differently, then scaling relationships calibrated on one architecture may not transfer directly to another. The comparative work discussed in the source review therefore supports architecture as a necessary control variable in scaling research \citep{wang2024,sengupta2025}.

\subsection{Systems Efficiency}

Systems efficiency targets the software and distributed-computing mechanisms that deliver model training. Large-scale training can incur substantial overhead from memory management, device communication, synchronization, and data movement. \citet{narayanan2021} demonstrated the importance of memory and compute optimization in large language-model training across GPU clusters. Quantization methods reduce representation precision and can lower memory pressure \citep{dettmers2023}. These interventions may not alter the theoretical token-to-parameter relationship, but they change the real compute cost of reaching a particular training configuration.

The distinction between theoretical and realized compute is important. A scaling strategy may be optimal in FLOP terms yet inefficient in wall-clock time or energy if the hardware is poorly utilized. Conversely, systems optimization can make a theoretically unchanged training objective substantially cheaper to execute. Resource-aware studies should therefore report both model-level and system-level measures whenever possible.

\subsection{Hardware and Edge Efficiency}

Hardware efficiency is the most direct expression of resource constraint. Instead of assuming access to large GPU clusters, edge and low-power research asks whether useful LLM training, adaptation, or inference can occur within restricted memory, energy, bandwidth, or processing budgets. This literature is increasingly important because it tests whether model-efficiency principles remain meaningful when hardware constraints dominate the optimization problem.

The five categories are summarized in Table~\ref{tab:taxonomy}.

\begin{table}[htbp]
\centering
\caption{Five Dimensions of Resource-Efficient LLM Research}
\label{tab:taxonomy}
\small
\begin{tabularx}{\textwidth}{@{}p{0.18\textwidth}Xp{0.28\textwidth}@{}}
\toprule
\textbf{Dimension} & \textbf{Primary optimization question} & \textbf{Representative literature from the source review} \\
\midrule
Data/token & How much and what kind of data are needed to obtain useful learning? & \citet{hoffmann2022}; \citet{sorscher2022}; \citet{pang2024} \\
Parameter & How much useful performance is obtained from model capacity and compute? & \citet{maloney2022}; \citet{boopathy2024}; \citet{hoffmann2022} \\
Architecture & Can model structure reduce active compute or improve scaling behavior? & \citet{tay2022}; \citet{wang2024}; \citet{sengupta2025} \\
Systems & Can memory, communication, synchronization, or numerical precision reduce realized cost? & \citet{narayanan2021}; \citet{dettmers2023} \\
Hardware/edge & Can LLMs be trained, adapted, or served under explicit memory, power, or compute constraints? & \citet{ahmed2025}; \citet{li2024}; \citet{yi2023}; \citet{yu2024}; \citet{moitra2025}; \citet{wei2025} \\
\bottomrule
\end{tabularx}
\end{table}

\section{Large Language Models in Resource-Constrained and Edge Environments}

Low-power and edge computing provide a stringent test of efficiency claims because they make resource limits unavoidable. Whereas frontier-scale training can compensate for inefficiency by adding hardware, constrained platforms require explicit trade-offs among memory, compute, precision, batch size, communication, and model capacity.

The source review identified several studies illustrating this transition. \citet{ahmed2025} examined cross-layer optimization on low-power devices, adjusting GPU frequency and batch size to accelerate training while reducing energy demand. \citet{li2024} proposed TPI-LLM for serving models at the 70B scale on low-resource edge devices through memory scheduling and communication algorithms. \citet{wei2025} discussed T-MAC, a low-bit approach aimed at efficient LLM inference on edge devices. \citet{moitra2025} introduced MEADOW, focusing on memory-efficient dataflow and data packing to reduce off-chip access. \citet{yu2024} proposed EDGE-LLM for efficient adaptation through compression and adaptive layer tuning, while \citet{yi2023} developed EdgeMoE for sparse LLM inference on mobile devices.

These studies demonstrate that constrained LLM operation is feasible, but they also reveal an important asymmetry in the literature. Most edge research focuses on engineering optimizations---quantization, memory scheduling, compression, dataflow, or inference engines---rather than directly testing the scaling relationships established by Kaplan and Hoffmann. The result is a gap between two mature ideas: scaling laws describe how model, data, and compute interact, while edge research describes how systems survive under limited resources. Much less work directly asks whether compute-optimal token-to-parameter principles remain valid when the entire experiment is moved into a constrained hardware regime.

This gap matters because the dominant scaling literature was developed using substantial compute. If scaling relationships depend on architecture, optimizer behavior, data quality, or hardware utilization, then results observed at enterprise scale may not generalize unchanged to a single-device environment. Resource-constrained experimentation should therefore be treated not only as an engineering problem, but also as a way to test the external validity of scaling theory.

\section{Training Cost, Accessibility, and Sustainability}

Efficiency also has an institutional dimension. The rising cost of frontier AI training can restrict meaningful experimentation to organizations capable of financing large compute budgets. \citet{cottier2024} documented the increasing cost trajectory of frontier-model training, while industry reporting summarized in the source review described individual training runs reaching hundreds of millions of dollars and potentially more \citep{hutong2024,shibu2024}. Even if precise figures vary across organizations and are difficult to verify, the directional problem is clear: the capital cost of frontier-scale experimentation creates barriers for smaller laboratories, startups, educational institutions, and independent researchers.

Environmental burden compounds the access problem. Data-center expansion associated with AI development has been linked in public reporting to substantial increases in greenhouse-gas emissions at major technology firms \citep{jackson2024}. Compute efficiency therefore affects more than model economics. A training strategy that obtains comparable performance from fewer resources can also reduce energy demand and the infrastructure required to support experimentation.

Policy discussions concerning AI transparency, accountability, and access respond indirectly to these pressures. The dissertation reviewed European Union and United States policy documents addressing risk, transparency, governance, and AI adoption. Those documents do not determine model architecture or scaling strategy, but they reflect the broader social consequences of concentrated and opaque AI development. For the present review, the most relevant point is narrower: inefficient scaling can contribute to economic and infrastructural concentration, while resource-efficient and reproducible approaches can lower barriers to participation.

Open models partially mitigate these barriers because they expose architectures, weights, code, or training artifacts that can be studied independently. Yet open development also brings maintenance and security challenges when projects depend on limited contributor resources \citep{openssf2024,openssf2025}. Accessibility should therefore not be equated with zero cost. Even when a model is openly available, researchers still need sufficient compute, storage, and engineering expertise to reproduce or extend results.

\section{Methodological Implications for Efficiency Research}

The scaling literature combines empirical and theoretical research. Empirical studies such as Kaplan and Hoffmann derive scaling relationships from repeated training observations across model and data configurations, while theoretical studies seek mathematical explanations for the regularities that emerge \citep{bahri2021,hutter2021,maloney2022}. Both approaches are necessary. Theory can explain why scaling behavior may arise, while experiments determine whether those relationships hold under real model architectures, datasets, and hardware.

The source review identified a methodological imbalance relevant to constrained-compute research. Many influential LLM studies rely on large-scale or one-shot experiments conducted on enterprise-grade infrastructure. Such work is valuable but difficult to reproduce for researchers without comparable hardware. Resource-constrained research offers a complementary methodological opportunity: smaller models can make repeated trials, controlled manipulations, and transparent logging more feasible.

Repeated experimentation is especially important because training outcomes contain stochastic variation. When researchers compare token counts, data-selection strategies, or efficiency interventions using a single run per condition, observed differences can be confounded by initialization, batch ordering, system behavior, or other uncontrolled factors. The dissertation literature review therefore emphasized repeated-measures designs as a methodological gap. Although repeated-measures terminology originates in statistical experimental design, the underlying principle is directly applicable to computational experiments: apply multiple conditions to the same standardized experimental unit, control as much configuration variance as possible, and repeat trials sufficiently to estimate variability rather than reporting a single deterministic-looking outcome \citep{creswell2022}.

Reproducibility is consequently part of compute efficiency. A result that requires inaccessible hardware or undocumented configuration cannot be meaningfully tested by much of the research community. The literature on machine-learning reproducibility has long emphasized the importance of code, configuration disclosure, repeated trials, and transparent reporting \citep{henderson2018,pineau2021,raff2019}. In LLM research, these practices are particularly important because even modest changes in hardware, precision, data pipeline, or model configuration can alter realized compute cost.

\section{Research Gaps and Agenda}

The reviewed literature suggests that the field has made substantial progress in understanding how model scale, data scale, and compute interact. However, the synthesis also reveals a set of unresolved problems that become more visible when the scaling question is moved from frontier infrastructure to constrained environments.

\subsection{Empirical Gap: Scaling Laws Under Constrained Compute}

The clearest empirical gap is the limited direct validation of scaling-law principles on low-power or otherwise resource-constrained platforms. Kaplan, Hoffmann, and many subsequent scaling studies rely on compute environments far beyond the resources available to many independent researchers. Edge-focused studies, by contrast, typically optimize memory, quantization, inference, communication, or device utilization rather than directly testing token-to-parameter scaling relationships.

A useful next step is therefore to treat constrained hardware as an experimental domain rather than merely a deployment target. Controlled experiments can vary training tokens while holding model architecture and hardware fixed, or vary model size while holding data and compute budgets constant. Such designs would test whether the shape of scaling relationships changes when hardware limits become binding.

\subsection{Theoretical Gap: No Standard Definition of Parameter Efficiency}

The literature uses the language of efficiency frequently but measures it inconsistently. Loss, perplexity, throughput, memory consumption, training time, parameter count, energy, and dollar cost can all serve as denominators or outcomes. Without a standard definition, ``more efficient'' can mean faster, smaller, cheaper, lower-loss, lower-memory, or lower-energy depending on the study.

A research agenda for parameter efficiency should therefore distinguish at least three levels: model-quality efficiency, compute efficiency, and energy efficiency. Equation~\ref{eq:pe} provides one possible model-quality-per-capacity-and-compute formulation, but additional work is required before any metric can be treated as general. A particularly important extension would incorporate power consumption so that performance can be expressed relative not only to parameters and floating-point throughput but also to energy demand.

\subsection{Methodological Gap: Repeated and Reproducible Efficiency Experiments}

The third gap concerns experimental design. Resource-efficient LLM research often compares methods across heterogeneous models, datasets, hardware, and evaluation metrics. These comparisons are informative but make causal interpretation difficult. More controlled studies are needed in which one efficiency variable is changed while the rest of the training environment remains fixed.

Repeated trials should also become more common in constrained-compute work. Smaller models and lower-cost platforms make this feasible. Repeated designs can estimate within-condition variability, reveal unstable configurations, and prevent a single favorable run from being mistaken for a general efficiency improvement. Public scripts, fixed environments, version-controlled configurations, and raw logs would further strengthen such studies.

\subsection{Data-Quality Gap: Tokens Are Not Interchangeable Units of Information}

Compute-optimal scaling treats token volume as a central variable, but data-pruning research demonstrates that the informational value of tokens is heterogeneous \citep{sorscher2022}. Two training corpora with the same number of tokens may differ in redundancy, diversity, noise, and relevance. As a result, future scaling studies should report not only the quantity of training tokens but also the method used to select, filter, deduplicate, or curate them.

This issue is particularly important for resource-constrained settings because better data may substitute for unavailable compute. When the hardware budget is fixed, improving the utility of each token may produce larger returns than increasing raw dataset size.

\subsection{Architecture and Systems Gap: Scaling Laws Do Not Operate Above the Hardware Stack}

Scaling relationships are often discussed at the level of abstract compute, but realized performance depends on architecture and systems implementation. Sparse activation, mixture-of-experts routing, quantization, memory scheduling, and communication overhead can all change the relationship between nominal FLOPs and actual training cost. Scaling studies therefore need clearer separation between theoretical compute and realized system cost.

Cross-platform replication would be especially valuable. If the same token-to-parameter configuration produces different efficiency behavior on a GPU cluster, a single accelerator, and an edge device, the difference would help identify where scaling theory ends and systems effects begin.

\subsection{Accessibility Gap: Efficiency as a Condition for Participation}

Finally, the literature suggests that compute efficiency should be treated as an access issue as well as a technical issue. The concentration of frontier-model training in well-funded organizations can limit independent validation and methodological diversity \citep{cottier2024}. Resource-conscious experimental frameworks may broaden participation by reducing the minimum hardware and financial requirements needed to test scaling hypotheses.

The objective is not to argue that small systems can replace frontier infrastructure. Rather, constrained environments can provide a complementary research layer: hypotheses can be tested cheaply, methods can be replicated more widely, and efficiency claims can be evaluated under conditions that better reflect the resources available to universities, small organizations, and independent researchers.

\section{Discussion: Scaling Down the Scaling Laws}

The central argument of this review is that scaling laws become more, not less, important as compute becomes constrained. Frontier-scale research asks how performance changes as resources increase. Resource-constrained research asks a related inverse question: given a fixed resource budget, which allocation of parameters, tokens, architecture, and systems effort produces the most useful outcome? Both questions are manifestations of the same optimization problem.

This perspective clarifies the progression of the literature. Kaplan established that scaling behavior is predictable. Hoffmann showed that predictable scaling does not justify indiscriminate model growth and that token-to-parameter balance matters under fixed compute. Data-centric, architectural, and systems research then expanded the optimization space by demonstrating that the utility of data, active parameters, precision, memory, and communication also affects realized efficiency. Edge research pushes the argument further by making hardware constraints explicit.

The phrase \emph{scaling down the scaling laws} therefore has two meanings. First, it refers to applying scaling principles to smaller models and smaller hardware environments. Second, it refers to decomposing the idea of ``scale'' into the resource dimensions that actually determine cost: tokens, parameters, active computation, memory, communication, energy, and time. A resource-constrained science of LLM training should measure these dimensions directly rather than treating model size as a sufficient proxy.

This synthesis also suggests that parameter efficiency can serve as a useful organizing concept even before the field converges on a single metric. Parameter efficiency asks whether model capacity is being converted into useful performance at a reasonable compute cost. That question connects compute-optimal training to data quality, architectural sparsity, systems optimization, and low-power hardware. The concept is therefore broader than parameter-efficient fine-tuning and narrower than generic ``AI efficiency'': it focuses attention on the productive use of model capacity under finite resources.

At the same time, the review cautions against a universal scaling prescription. The literature itself identifies architecture-dependent, data-dependent, and context-dependent limits to simple power-law extrapolation \citep{sengupta2025,wang2024,sorscher2022}. Compute-optimal behavior should therefore be treated as an empirical relationship to be tested in each regime, not as a guarantee that one token-to-parameter rule applies across all models and hardware.

\section{Conclusion}

LLM research has moved from a period dominated by scale maximization toward a more explicit concern with resource allocation. Empirical scaling laws established that model performance changes predictably as parameters, data, and compute increase, but those same laws also exposed diminishing returns. Compute-optimal training reframed the problem by showing that the relationship between training tokens and model parameters matters under finite compute. Subsequent work on data pruning, sparse architectures, systems optimization, quantization, and edge computing has expanded the efficiency problem beyond model size alone.

The resulting literature supports three conclusions. First, efficient LLM training is multidimensional: data, parameters, architecture, systems, and hardware interact. Second, resource-constrained environments are not merely deployment targets; they are important experimental settings for testing whether scaling principles generalize beyond enterprise infrastructure. Third, the field needs more consistent efficiency metrics and more reproducible experimental designs if claims about ``efficient'' models are to be compared meaningfully.

Future research should therefore move in two complementary directions. Scaling theory should be tested downward through controlled experiments on smaller models and constrained hardware, while efficiency measurement should become more comprehensive by incorporating model quality, compute, and eventually energy use. The long-term objective is not to abandon scaling, but to make scaling conditional on resource effectiveness. In that sense, the next phase of LLM research may depend less on how much compute can be accumulated than on how intelligently available compute can be used.

\section*{Acknowledgment of Dissertation Origin}

This manuscript is derived from and substantially refactors the literature review chapter of the author's doctoral dissertation. The article reorganizes the dissertation material as a standalone scholarly synthesis focused on scaling laws, parameter efficiency, compute-optimal training, and resource-constrained LLM research. A later revision will update the literature search to a clearly stated publication cutoff date and reverify all comparative model statistics against primary sources.


\begin{thebibliography}{99}
\raggedright
\setlength{\emergencystretch}{2em}
\Urlmuskip=0mu plus 1mu\relax

\bibitem[Ahmed et al.(2025)]{ahmed2025}
Ahmed, M. A., Pfeiffer, K., Khdr, H., Abboud, O., Khalili, R., \& Henkel, J. (2025).
Accelerated training on low-power edge devices.
\textit{arXiv preprint arXiv:2502.18323}. \url{https://arxiv.org/abs/2502.18323}

\bibitem[Bahri et al.(2021)]{bahri2021}
Bahri, Y., Dyer, E., Kaplan, J., Lee, J., \& Sharma, U. (2021).
Explaining neural scaling laws.
\textit{arXiv preprint arXiv:2102.06701}. \url{https://arxiv.org/abs/2102.06701}

\bibitem[Bernardelle and Demartini(2024)]{bernardelle2024}
Bernardelle, P., \& Demartini, G. (2024).
Optimizing LLMs with direct preferences: A data efficiency perspective.
\textit{Proceedings of SIGIR-AP 2024}, 236--240.
\url{https://doi.org/10.1145/3673791.3698411}

\bibitem[Boopathy and Fiete(2024)]{boopathy2024}
Boopathy, A., \& Fiete, I. (2024).
Unified neural network scaling laws and scale-time equivalence.
\textit{arXiv preprint}. \url{https://arxiv.org/abs/2409.05782}

\bibitem[Brown et al.(2020)]{brown2020}
Brown, T. B., Mann, B., Ryder, N., et al. (2020).
Language models are few-shot learners.
\textit{arXiv preprint}. \url{https://arxiv.org/abs/2005.14165}

\bibitem[Javaheripi and Bubeck(2023)]{javaheripi2023}
Javaheripi, M., \& Bubeck, S. (2023, December 12).
Phi-2: The surprising power of small language models.
\textit{Microsoft Research Blog}.
\url{https://www.microsoft.com/en-us/research/blog/phi-2-the-surprising-power-of-small-language-models/}

\bibitem[Cambria and White(2014)]{cambria2014}
Cambria, E., \& White, B. (2014).
Jumping NLP curves: A review of natural language processing research.
\textit{IEEE Computational Intelligence Magazine, 9}(2), 48--57.
\url{https://doi.org/10.1109/MCI.2014.2307227}

\bibitem[Chowdhery et al.(2022)]{chowdhery2022}
Chowdhery, A., Narang, S., Devlin, J., et al. (2022).
PaLM: Scaling language modeling with pathways.
\textit{arXiv preprint}. \url{https://arxiv.org/abs/2204.02311}

\bibitem[Cottier et al.(2024)]{cottier2024}
Cottier, B., Rahman, R., Fattorini, L., Maslej, N., Besiroglu, T., \& Owen, D. (2024).
The rising costs of training frontier AI models.
\textit{arXiv preprint arXiv:2405.21015}. \url{https://arxiv.org/abs/2405.21015}

\bibitem[Creswell and Creswell(2022)]{creswell2022}
Creswell, J. W., \& Creswell, J. D. (2022).
\textit{Research design: Qualitative, quantitative, and mixed methods approaches} (6th ed.).
SAGE.

\bibitem[Dettmers et al.(2023)]{dettmers2023}
Dettmers, T., Pagnoni, A., Holtzman, A., \& Zettlemoyer, L. (2023).
QLoRA: Efficient finetuning of quantized LLMs.
\textit{Advances in Neural Information Processing Systems, 36}.
\url{https://arxiv.org/abs/2305.14314}

\bibitem[Devlin et al.(2019)]{devlin2019}
Devlin, J., Chang, M.-W., Lee, K., \& Toutanova, K. (2019).
BERT: Pre-training of deep bidirectional transformers for language understanding.
\textit{Proceedings of NAACL-HLT}, 4171--4186.
\url{https://doi.org/10.18653/v1/N19-1423}

\bibitem[Goldberg(2016)]{goldberg2016}
Goldberg, Y. (2016).
A primer on neural network models for natural language processing.
\textit{Journal of Artificial Intelligence Research, 57}, 345--420.
\url{https://doi.org/10.1613/jair.4992}

\bibitem[Gunasekar et al.(2023)]{gunasekar2023}
Gunasekar, S., Zhang, Y., Aneja, J., et al. (2023).
Textbooks are all you need.
\textit{arXiv preprint arXiv:2306.11644}. \url{https://arxiv.org/abs/2306.11644}

\bibitem[Henderson et al.(2018)]{henderson2018}
Henderson, P., Islam, R., Bachman, P., Pineau, J., Precup, D., \& Meger, D. (2018).
Deep reinforcement learning that matters.
\textit{Proceedings of the AAAI Conference on Artificial Intelligence, 32}(1).
\url{https://arxiv.org/abs/1709.06560}

\bibitem[Hoffmann et al.(2022)]{hoffmann2022}
Hoffmann, J., Borgeaud, S., Mensch, A., et al. (2022).
Training compute-optimal large language models.
\textit{arXiv preprint arXiv:2203.15556}. \url{https://arxiv.org/abs/2203.15556}

\bibitem[Hu and Tong(2024)]{hutong2024}
Hu, K., \& Tong, A. (2024, November 11).
OpenAI and rivals seek new path to smarter AI as current methods hit limitations.
\textit{Reuters}.
\href{https://www.reuters.com/technology/artificial-intelligence/openai-rivals-seek-new-path-smarter-ai-current-methods-hit-limitations-2024-11-11/}{Reuters article}

\bibitem[Hutter(2021)]{hutter2021}
Hutter, M. (2021).
Learning curve theory.
\textit{arXiv preprint arXiv:2102.04074}. \url{https://arxiv.org/abs/2102.04074}

\bibitem[Jackson(2024)]{jackson2024}
Jackson, F. (2024, July 5).
Google's greenhouse gas emissions increased by 48\% since 2019, thanks to AI pursuits.
\textit{TechRepublic}.
\url{https://www.techrepublic.com/article/google-ai-environmental-impact/}

\bibitem[Jiang et al.(2023)]{jiang2023}
Jiang, A. Q., Sablayrolles, A., Mensch, A., Bamford, C., Chaplot, D. S., de las Casas, D., Bressand, F., Lengyel, G., Lample, G., Saulnier, L., Renard Lavaud, L., Lachaux, M.-A., Stock, P., Le Scao, T., Lavril, T., Wang, T., Lacroix, T., \& El Sayed, W. (2023).
Mistral 7B.
\textit{arXiv preprint arXiv:2310.06825}. \url{https://arxiv.org/abs/2310.06825}

\bibitem[Kaplan et al.(2020)]{kaplan2020}
Kaplan, J., McCandlish, S., Henighan, T., Brown, T. B., Chess, B., Child, R., Gray, S., Radford, A., Wu, J., \& Amodei, D. (2020).
Scaling laws for neural language models.
\textit{arXiv preprint arXiv:2001.08361}. \url{https://arxiv.org/abs/2001.08361}

\bibitem[Knuth(1997)]{knuth1997}
Knuth, D. E. (1997).
\textit{The art of computer programming, Volume 1: Fundamental algorithms} (3rd ed.).
Addison-Wesley.

\bibitem[Li et al.(2024)]{li2024}
Li, Z., Feng, W., Guizani, M., \& Yu, H. (2024).
TPI-LLM: Serving 70B-scale LLMs efficiently on low-resource edge devices.
\textit{arXiv preprint arXiv:2410.00531}. \url{https://arxiv.org/abs/2410.00531}

\bibitem[Lieber et al.(2021)]{lieber2021}
Lieber, O., Sharir, O., Lenz, B., \& Shoham, Y. (2021).
Jurassic-1: Technical details and evaluation.
\textit{AI21 Labs White Paper}.
\url{https://sharir.org/papers/jurassic_white_paper.pdf}

\bibitem[Maloney et al.(2022)]{maloney2022}
Maloney, A., Roberts, D. A., \& Sully, J. (2022).
A solvable model of neural scaling laws.
\textit{arXiv preprint}. \url{https://arxiv.org/abs/2210.16859}

\bibitem[Mele and Laccetti(2023)]{mele2023}
Mele, V., \& Laccetti, G. (2023).
Algorithm and software overhead: A theoretical approach to performance portability.
In R. Wyrzykowski, E. Deelman, J. Dongarra, \& K. Karczewski (Eds.), \textit{Parallel processing and applied mathematics} (Vol. 13827, pp. 89--100). Springer.
\url{https://doi.org/10.1007/978-3-031-30445-3_8}

\bibitem[Gemma Team et al.(2024)]{mesnard2024}
Gemma Team, Mesnard, T., Hardin, C., Dadashi, R., Bhupatiraju, S., Pathak, S., Sifre, L., et al. (2024).
Gemma: Open models based on Gemini research and technology.
\textit{arXiv preprint arXiv:2403.08295}. \url{https://arxiv.org/abs/2403.08295}

\bibitem[Mikolov et al.(2013)]{mikolov2013}
Mikolov, T., Chen, K., Corrado, G., \& Dean, J. (2013).
Efficient estimation of word representations in vector space.
\textit{arXiv preprint arXiv:1301.3781}. \url{https://arxiv.org/abs/1301.3781}

\bibitem[Moitra et al.(2025)]{moitra2025}
Moitra, A., Ghosh, A., Agarwal, S., Amarnath, A., Swaminathan, K., \& Panda, P. (2025).
MEADOW: Memory-efficient dataflow and data packing for low power edge LLMs.
\textit{arXiv preprint arXiv:2503.11663}. \url{https://doi.org/10.48550/arXiv.2503.11663}

\bibitem[Narayanan et al.(2021)]{narayanan2021}
Narayanan, D., Shoeybi, M., Casper, J., LeGresley, P., Patwary, M., Korthikanti, V. A., Vainbrand, D., Kashinkunti, P., Bernauer, J., Catanzaro, B., Phanishayee, A., \& Zaharia, M. (2021).
Efficient large-scale language model training on GPU clusters using Megatron-LM.
\textit{Proceedings of the International Conference for High Performance Computing, Networking, Storage and Analysis (SC 2021)}.
\url{https://arxiv.org/abs/2104.04473}

\bibitem[OpenSSF(2024)]{openssf2024}
Open Source Security Foundation. (2024, December 4).
Open source usage trends and security challenges revealed in new study.
\textit{OpenSSF}.
\url{https://openssf.org/press-release/2024/12/04/open-source-usage-trends-and-security-challenges-revealed-in-new-study/}

\bibitem[OpenSSF(2025)]{openssf2025}
Open Source Security Foundation. (2025, January 23).
Predictions for open source security in 2025: AI, state actors, and supply chains.
\textit{OpenSSF}.
\url{https://openssf.org/blog/2025/01/23/predictions-for-open-source-security-in-2025-ai-state-actors-and-supply-chains/}

\bibitem[Pang et al.(2024)]{pang2024}
Pang, J., Wei, J., Shah, A. P., Zhu, Z., Wang, Y., Qian, C., Liu, Y., Bao, Y., \& Wei, W. (2024).
Improving data efficiency via curating LLM-driven rating systems.
\textit{arXiv preprint}. \url{https://doi.org/10.48550/arXiv.2410.10877}

\bibitem[Pennington et al.(2014)]{pennington2014}
Pennington, J., Socher, R., \& Manning, C. D. (2014).
GloVe: Global vectors for word representation.
\textit{Proceedings of the 2014 Conference on Empirical Methods in Natural Language Processing (EMNLP)}, 1532--1543.
\url{https://doi.org/10.3115/v1/D14-1162}

\bibitem[Peters et al.(2018)]{peters2018}
Peters, M. E., Neumann, M., Iyyer, M., Gardner, M., Clark, C., Lee, K., \& Zettlemoyer, L. (2018).
Deep contextualized word representations.
\textit{Proceedings of NAACL-HLT}, 2227--2237.
\url{https://doi.org/10.18653/v1/N18-1202}

\bibitem[Pineau et al.(2021)]{pineau2021}
Pineau, J., Vincent-Lamarre, P., Sinha, K., Larivi\`ere, V., Beygelzimer, A., d'Alch\'e-Buc, F., Fox, E., \& Larochelle, H. (2021).
Improving reproducibility in machine learning research: A report from the NeurIPS 2019 reproducibility program.
\textit{Journal of Machine Learning Research, 22}(164), 1--20.
\url{https://jmlr.org/papers/v22/20-303.html}

\bibitem[Radford et al.(2018)]{radford2018}
Radford, A., Narasimhan, K., Salimans, T., \& Sutskever, I. (2018).
Improving language understanding by generative pre-training.
\textit{OpenAI}.
\url{https://cdn.openai.com/research-covers/language-unsupervised/language_understanding_paper.pdf}

\bibitem[Radford et al.(2019)]{radford2019}
Radford, A., Wu, J., Child, R., Luan, D., Amodei, D., \& Sutskever, I. (2019).
Language models are unsupervised multitask learners.
\textit{OpenAI}.
\url{https://cdn.openai.com/better-language-models/language_models_are_unsupervised_multitask_learners.pdf}

\bibitem[Rae et al.(2021)]{rae2021}
Rae, J. W., Borgeaud, S., Cai, T., et al. (2021).
Scaling language models: Methods, analysis \& insights from training Gopher.
\textit{arXiv preprint}. \url{https://arxiv.org/abs/2112.11446}

\bibitem[Raff(2019)]{raff2019}
Raff, E. (2019).
A step toward quantifying independently reproducible machine learning research.
\textit{Advances in Neural Information Processing Systems, 32 (NeurIPS 2019)}.
\url{https://proceedings.neurips.cc/paper/2019/hash/c429429bf1f2af051f2021dc92a8ebea-Abstract.html}

\bibitem[Russell and Norvig(2021)]{russell2021}
Russell, S., \& Norvig, P. (2021).
\textit{Artificial intelligence: A modern approach} (4th ed.). Pearson.

\bibitem[Sengupta et al.(2025)]{sengupta2025}
Sengupta, A., Goel, Y., \& Chakraborty, T. (2025).
How to upscale neural networks with scaling law? A survey and practical guidelines.
\textit{arXiv preprint}. \url{https://arxiv.org/abs/2502.12051}

\bibitem[Shibu(2024)]{shibu2024}
Shibu, S. (2024, July 8).
How much does it cost to develop and train AI? Here's the current price, according to the CEO of an \$18 billion AI startup.
\textit{Entrepreneur}.
\url{https://www.entrepreneur.com/business-news/anthropic-ceo-ai-will-cost-10-billion-to-train-by-2025/476750}

\bibitem[Smith et al.(2022)]{smith2022}
Smith, S., Patwary, M., Norick, B., et al. (2022).
Using DeepSpeed and Megatron to train Megatron-Turing NLG 530B, a large-scale generative language model.
\textit{arXiv preprint}. \url{https://arxiv.org/abs/2201.11990}

\bibitem[Sorscher et al.(2022)]{sorscher2022}
Sorscher, B., Geirhos, R., Shekhar, S., Ganguli, S., \& Morcos, A. S. (2022).
Beyond neural scaling laws: Beating power law scaling via data pruning.
\textit{arXiv preprint}. \url{https://arxiv.org/abs/2206.14486}

\bibitem[Su et al.(2019)]{su2019}
Su, P., Wen, S., Yang, H., Chabbi, M., \& Liu, X. (2019).
Redundant loads: A software inefficiency indicator.
\textit{2019 IEEE/ACM 41st International Conference on Software Engineering (ICSE)}, 982--993.
\url{https://doi.org/10.1109/ICSE.2019.00103}

\bibitem[Wei et al.(2025)]{wei2025}
Wei, J., Cao, S., Cao, T., Ma, L., Wang, L., Zhang, Y., \& Yang, M. (2025).
T-MAC: CPU renaissance via table lookup for low-bit LLM deployment on edge.
\textit{Proceedings of the 2025 ACM European Conference on Computer Systems (EuroSys 2025)}, 278--292.
\url{https://doi.org/10.1145/3689031.3696099}

\bibitem[Tay et al.(2022)]{tay2022}
Tay, Y., Dehghani, M., Bahri, D., \& Metzler, D. (2022).
Efficient transformers: A survey.
\textit{ACM Computing Surveys, 55}(6), 1--28.
\url{https://doi.org/10.1145/3530811}

\bibitem[Vaswani et al.(2017)]{vaswani2017}
Vaswani, A., Shazeer, N., Parmar, N., et al. (2017).
Attention is all you need.
\textit{Advances in Neural Information Processing Systems, 30}, 5998--6008.
\url{https://doi.org/10.48550/arXiv.1706.03762}

\bibitem[Wang et al.(2024)]{wang2024}
Wang, S., Chen, Z., Li, B., He, K., Zhang, M., \& Wang, J. (2024).
Scaling laws across model architectures: A comparative analysis of dense and MoE models in large language models.
\textit{arXiv preprint arXiv:2410.05661}. \url{https://arxiv.org/abs/2410.05661}

\bibitem[Weizenbaum(1966)]{weizenbaum1966}
Weizenbaum, J. (1966).
ELIZA---A computer program for the study of natural language communication between man and machine.
\textit{Communications of the ACM, 9}(1), 36--45.
\url{https://doi.org/10.1145/365153.365168}

\bibitem[Yi et al.(2023)]{yi2023}
Yi, R., Guo, L., Wei, S., Zhou, A., Wang, S., \& Xu, M. (2023).
EdgeMoE: Empowering sparse large language models on mobile devices.
\textit{arXiv preprint arXiv:2308.14352}. \url{https://doi.org/10.48550/arXiv.2308.14352}

\bibitem[Yu et al.(2024)]{yu2024}
Yu, Z., Wang, Z., Li, Y., et al. (2024).
EDGE-LLM: Enabling efficient large language model adaptation on edge devices via layerwise unified compression and adaptive layer tuning and voting.
\textit{arXiv preprint arXiv:2406.15758}. \url{https://doi.org/10.48550/arXiv.2406.15758}

\bibitem[Zhang et al.(2024)]{zhang2024}
Zhang, P., Zeng, G., Wang, T., \& Lu, W. (2024).
TinyLlama: An open-source small language model.
\textit{arXiv preprint arXiv:2401.02385}. \url{https://arxiv.org/abs/2401.02385}

\end{thebibliography}
\end{document}